\documentclass[letterpaper]{article}
\usepackage{proceed2e}
\usepackage[margin=1in]{geometry}
\usepackage{hyperref}
\usepackage[linesnumbered,lined,commentsnumbered]{algorithm2e}
\usepackage{mathabx}
\usepackage{graphicx}
\usepackage[font=small,labelfont=bf]{caption}

\usepackage{times}

\title{Deep Contextual Multi-armed Bandits}

\author{ {\bf Mark Collier } \\
HubSpot, Inc. \\
Dublin, Ireland \\
\And
{\bf Hector Urdiales Llorens}  \\
HubSpot, Inc. \\
Dublin, Ireland \\
}

\begin{document}

\maketitle

\begin{abstract}
Contextual multi-armed bandit problems arise frequently in important industrial applications. Existing solutions model the context either linearly, which enables uncertainty driven (principled) exploration, or non-linearly, by using epsilon-greedy exploration policies. Here we present a deep learning framework for contextual multi-armed bandits that is both non-linear and enables principled exploration at the same time. We tackle the exploration vs.\ exploitation trade-off through Thompson sampling by exploiting the connection between inference time dropout and sampling from the posterior over the weights of a Bayesian neural network \cite{gal2016dropout}. In order to adjust the level of exploration automatically as more data is made available to the model, the dropout rate is learned rather than considered a hyperparameter. We demonstrate that our approach substantially reduces regret on two tasks (the UCI Mushroom task and the Casino Parity task) when compared to 1) non-contextual bandits, 2) epsilon-greedy deep contextual bandits, and 3) fixed dropout rate deep contextual bandits. Our approach is currently being applied to marketing optimization problems at HubSpot.
\end{abstract}

\section{INTRODUCTION}

The contextual bandit problem is a variant of the extensively studied multi-armed bandit problem \cite{auer2002finite}. Both contextual and non-contextual bandits involve making a sequence of decisions on which action to take from an action space $A$. After an action is taken, a stochastic reward $r$ is revealed for the chosen action only. The goal is to maximize these rewards. Maximization of reward is often recast as the minimization of regret \cite{bubeck2012regret}. As the reward is only revealed for the chosen action, bandit problems involve trading off exploration to try potentially new and better actions and exploitation of known good actions.

In a contextual bandit, before making each decision the bandit is shown some context $\mathbf{x} \in X$. Asymptotically optimal approaches to trading off exploration and exploitation for the non-contextual multi-armed bandit can be derived \cite{auer2002nonstochastic}. But deriving such solutions in the contextual case has resulted in simplifying assumptions such as assuming a linear relationship between the context and the reward \cite{li2010contextual}. These simplifying assumptions are often unsatisfactory.

For example, at HubSpot we wish to optimize the time of day to send an email in order to maximize the probability of an email converting (opened, clicked on or replied to). We know that the context of the particular email being sent, such as the timezone of the email sender, has an effect on whether the email converts for a given send time. Additionally, we only observe whether the email converts for the time we actually \textit{choose} to send the email. Thus, we model this as a contextual bandit problem. The action space $A$ is the possible times to send an email, the context $X$ are the features describing the email being sent and the reward $r$ is whether the email converts. We know from other non-bandit (classification) tasks involving the same email data, such as predicting the bounce rate of an email, that deep neural networks substantially outperform linear models at modeling our email data.

Many other marketing problems can also be cast as contextual bandit problems \cite{lu2010contextual} and we wish to model the context of these problems using deep neural networks, while still managing the exploration vs.\ exploitation trade-off. We propose doing so by applying recent advances in Bayesian neural networks, whereby it is possible to obtain samples from an approximate posterior over the weights in a neural network. Given these samples it is trivial to use Thompson sampling \cite{thompson1933likelihood} to manage the exploration vs.\ exploitation trade-off.


\section{BACKGROUND}


\subsection{CONTEXTUAL MULTI-ARMED BANDITS}

We provide a more formal definition of the contextual bandit problem. Suppose we have an agent acting in an environment. At each timestep the agent is presented with some context $\mathbf{x} \in X$ from the environment. The agent must choose to take some action $\mathbf{a} \in A$ from a set of possible actions \{$\mathbf{a_1}$, $\mathbf{a_2}$, ..., $\mathbf{a_m}$\}. When the agent takes an action it receives a real-valued reward $r$ for the action taken, however it does not observe a reward for untaken actions. For simplicity we restrict $r \in \{0,1\}$ although the treatment is trivially extended to any real valued reward. The agent wishes to maximize the cumulative reward over time, and in our formulation the agent is assumed to interact with the environment over an infinite time horizon.

The agent must build a model for $P(r | \mathbf{a}, \mathbf{x})$ from observed ($\mathbf{x}$, $\mathbf{a}$, r) triplets. But the agent only observes triplets involving actions $\mathbf{a}$ it has taken, so the agent drives its own data generation process. Thus, it is necessary for the agent to trade off exploration and exploitation. If the agent learns early on to associate high rewards with a particular action, it may not take actions which have higher expected reward in new, unseen contexts. Equally, the agent cannot always take random actions as it will forgo higher expected reward for known good actions in particular contexts.

There are many approaches to trading off exploration and exploitation \cite{auer2002using,auer2002finite,silver2016mastering}. Two popular and simple approaches are epsilon-greedy exploration and Thompson sampling. With an epsilon-greedy policy, a random action is taken $\epsilon$ percentage of the time and the best predicted action $1 - \epsilon$ percentage of the time. 

Thomson sampling \cite{thompson1933likelihood} is a heuristic for trading off exploration and exploitation when using a parametric likelihood $P(r | \mathbf{a}, \mathbf{x}; \mathbf{w})$ function and a posterior on the parameters $P(\mathbf{w} | \{(\mathbf{x}, \mathbf{a}, r)\})$. At each timestep we sample $\mathbf{\tilde{w}} \sim P(\mathbf{w} | \{(\mathbf{x}, \mathbf{a}, r)\})$ and then choose the action with highest expected reward $P(r | \mathbf{a}, \mathbf{x}; \mathbf{\tilde{w}})$. Thompson sampling has been shown to be a very effective heuristic for managing this trade-off \cite{agrawal2013thompson,chapelle2011empirical}.

\subsection{BAYESIAN NEURAL NETWORKS}

Neural networks are typically trained with maximum likelihood estimation (MLE) or maximum a posteriori estimation (MAP) of the network weights \cite{blundell2015weight}. This leads to a point estimate of the weights and thus a Bayesian approach to ascertaining uncertainty estimates from the network is not possible. In contrast, in a Bayesian neural network a prior is placed on the network weights and the posterior distribution is learned over these weights. In general due to the intractability of full Bayesian inference one conducts approximate Bayesian inference to learn the posterior.

Approximate inference in Bayesian neural networks has a long history 
\cite{barber1998ensemble,buntine1991bayesian,hinton1993keeping,mackay1992practical,mackay1995probable,neal1995bayesian}. Only recently have approaches to approximate inference been proposed which scale well to larger networks and datasets \cite{blundell2015weight,gal2016dropout,graves2011practical}.

Of particular importance to our application is a recent result \cite{gal2016dropout} where dropout training \cite{srivastava2014dropout} in an arbitrary depth feed-forward neural network with arbitrary non-linearities is shown to be equivalent to an approximation of the probabilistic deep Gaussian process \cite{damianou2013deep}. In practice this Bayesian interpretation of dropout means that if we turn dropout on at inference time, each stochastic forward pass through the network corresponds to a sample from an approximate posterior over the network weights. This property of dropout at inference time has been used to undertake Thompson sampling to drive exploration in reinforcement learning problems \cite{gal2016dropout}.

Under the Bayesian interpretation of dropout, a continuous relaxation of Bernoulli dropout using the Concrete distribution \cite{maddison2016concrete} can be realized \cite{gal2017concrete}. This continuous relaxation of dropout, named Concrete Dropout, treats the dropout parameter $p$ (the probability of turning off a node in Bernoulli dropout) as a parameter of the network that can be trained using standard gradient descent methods. Appropriate regularization parameters are then added to the objective function, which trades off between fitting the data and maximizing the entropy of the dropout parameter. As the amount of data increases, the data likelihood term overwhelms the dropout regularization term, which leads to well calibrated uncertainty estimates, and thus appropriate levels of exploration.

\section{DEEP CONTEXTUAL MULTI-ARMED BANDITS}

In our model, we use a neural network to model $P(r | \mathbf{a}, \mathbf{x}; \mathbf{w})$, where $\mathbf{w}$ are the network weights. We train the network using Concrete Dropout, and do not disable dropout at inference time.

At each timestep we are presented with a context $\mathbf{x}$ and a set of possible actions \{$\mathbf{a_1}$, $\mathbf{a_2}$, ..., $\mathbf{a_m}$\}. We sample dropout noise $\mathbf{d}$ from a uniform distribution, $U(0, 1)$, which is used to, in effect, sample from the posterior over the weights $\mathbf{w}$. We unroll the $m$ actions into  ($\mathbf{x}$, $\mathbf{a_i}$) pairs which are passed to the network for inference. We choose the action $\mathbf{a}$ whose ($\mathbf{x}$, $\mathbf{a_i}$) pair has the highest expected reward. Theoretically for Thompson sampling we should update the model after each observation. In practice, realtime online learning is unsuitable for industrial settings, so we retrain the model on an exponential scale in the number of data points. See \autoref{algo:dcmab}.

\IncMargin{1em}
\begin{algorithm}
 \SetKwInOut{Input}{input}\SetKwInOut{Output}{output}
 \Input{$N$ the number of initial steps to take actions randomly before training, $K$ coefficient for retraining on an exponential scale and a neural network architecture that defines a parametric function $f(\mathbf{x}, \mathbf{a}; \mathbf{w})$.}
 $nextRetrain\leftarrow N$\;
 \While{True}{
  \For{$i\leftarrow 1$ \KwTo $nextRetrain$}{
   receive context $\mathbf{x}$ and possible actions \{$\mathbf{a_1}$, $\mathbf{a_2}$, ..., $\mathbf{a_m}$\}\;
   sample dropout noise $\mathbf{d}$\;
   compute corresponding weights $\mathbf{w}_d$\;
   \For{$\mathbf{a} \in \{\mathbf{a_1}, \mathbf{a_2}, ..., \mathbf{a_m}\}$}{
   compute predicted reward $r = f(\mathbf{x}, \mathbf{a}; \mathbf{w}_d)$\;
   }
   choose $\mathbf{a}$ with the highest predicted reward\;
  }
  retrain on all ($\mathbf{x}$, $\mathbf{a}$, r) triplets\;
  $nextRetrain\leftarrow K*nextRetrain$\;
 }
 \caption{Deep Contextual Multi-Armed Bandits\label{algo:dcmab}}
\end{algorithm}
\DecMargin{1em}

Initially, when the amount of data is small, the uncertainty on the network weights is high. Therefore, samples from the posterior have high variance, which leads to a high level of exploration, as desired. As the model observes more ($\mathbf{x}$, $\mathbf{a}$, r) triplets, the uncertainty on the weights decreases, and the network explores less and less, converging to an almost purely exploitative strategy. Additionally, by using Concrete Dropout, the level of exploration is automatically tuned according to the size of the network, the amount of data observed, and the complexity of the context. Thus, we do not have to create hand-crafted dropout rate or epsilon annealing schedules for each problem. This is particularly important in real world settings, where it may not be possible to run a large number of simulations to determine an appropriate problem-specific annealing schedule.

In summary, our approach satisfies three key requirements:

\begin{enumerate}
\setlength\itemsep{0em}
  \item It models the context $\mathbf{x}$ using a deep non-linear neural network.
  \item It explores the $X \bigtimes A$ space in a principled and efficient way that requires little manual tuning.
  \item Its time complexity is consistent with real-world, latency-constrained tasks.
\end{enumerate}

\section{EXPERIMENTS}

We test our approach on two tasks with synthetic data which allows us to run many simulations. We compare against three baseline approaches: 1) a non-contextual bandit which uses Thompson sampling under a Beta/Binomial conjugate Bayesian analysis, 2) an epsilon-greedy bandit which chooses a random action 5\% of the time and the best possible action the remainder of the time and 3) a bandit with fixed dropout rate.

In the below experiments for the contextual models (including the epsilon-greedy and fixed dropout rate bandits), we use a neural network with 2 hidden layers with 256 units each and ReLU activation function. All networks are trained using the Adam optimizer \cite{kingma2014adam} with initial learning rate = 0.001. We retrain the contextual model on an exponential scale after observing $2^7$, $2^8$, $2^9$, ... examples i.e. $K = 2$ and $N = 128$.

\subsection{MUSHROOM TASK}

We make a slight modification to a previously used contextual bandit problem \cite{blundell2015weight,guez2015sample}. The UCI Mushroom dataset \cite{UCI2013} contains 8,124 mushrooms classified as poisonous or edible, with up to 22 features for each mushroom. At each timestep, our agent is presented with the features for a mushroom and must decide whether to eat it or not. The agent gets a reward of 1 for eating an edible mushroom, a reward of 0 for eating a poisonous mushroom, and with probability $p$ a reward of 1 for not eating a mushroom. In the below experiments we set $p = 0.5$. Regret is measured against an oracle which can see whether the mushroom is edible or not, always eats edible mushrooms and does not eat poisonous mushrooms.

\begin{figure}[h]
\setlength{\belowcaptionskip}{-10pt}
\centering
\includegraphics[width=0.4\textwidth]{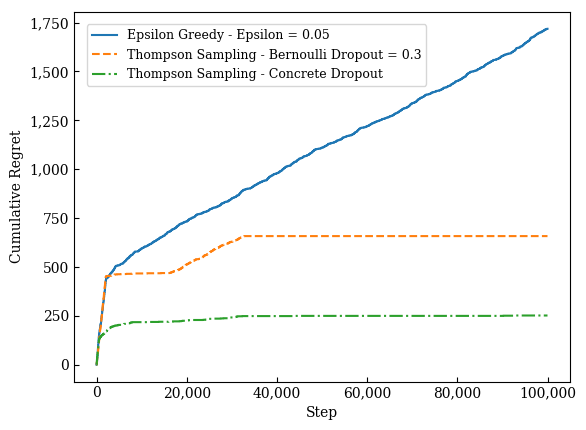}
\caption{Mushroom Task - comparison between epsilon-greedy, Thompson sampling with Bernoulli dropout, and Thompson sampling with Concrete Dropout. Concrete Dropout with final cumulative regret (FCR) of 252 performs significantly better than Bernoulli dropout which has FCR of 658 while the FCR for for the epsilon-greedy agent is 1,718.}
\label{fig:mushroom}
\end{figure}

\autoref{fig:mushroom} demonstrates that, as expected, the epsilon-greedy agent accumulates regret linearly, due to taking random actions 5\% of the time. Both fixed rate dropout and Concrete Dropout agents rapidly learn to ``solve'' the task, but Concrete Dropout not only has a much lower final cumulative regret than the fixed rate dropout agent (less than half), but it also learns to stop exploring earlier. Not shown is a non-contextual bandit whose final regret is 24,048, which demonstrates the value of modeling the context for this task. The discontinuities in the graph are due to the fact that we retrain on an exponential scale.

\subsection{CASINO PARITY TASK}

We propose a new contextual bandit task which is a twist on the traditional non-contextual multi-armed bandit problem. Suppose there are L bandits in a casino, each of type A or B. Bandits of type A pay out with probability $p_A$ and bandits of type B pay out with probability $p_B$. Assume for simplicity the payouts are always one dollar. Every bandit in the casino has an ID encoded as a binary string. Unknown to the gambler, bandits of type A have an even parity ID and bandits of type B an odd parity ID. At each timestep the gambler is presented the ID of a bandit of either type A or B, and the gambler must choose to play or not. If the gambler plays they receive reward according to the bandit type's probability, otherwise they are randomly assigned one of the L bandits which they must play. Without loss of generality if $p_A > p_B$, clearly the optimal strategy is to always play the even parity bandits and to not play any odd parity bandits. Rewards are stochastic and the context requires a non-linear model \cite{minsky1969perceptrons}. When $p_A \neq p_B$ the context can be leveraged to minimize regret.

In the below experiments we set $p_A = 0.7$, $p_A = 0.3$ and $L = 32$. Regret is measured relative to an oracle which can see the bandit type A or B and always plays type A bandits and always refuses to play type B bandits if $p_A > p_B$ and vice versa.

\begin{figure}[h]
\setlength{\belowcaptionskip}{-10pt}
\centering
\includegraphics[width=0.4\textwidth]{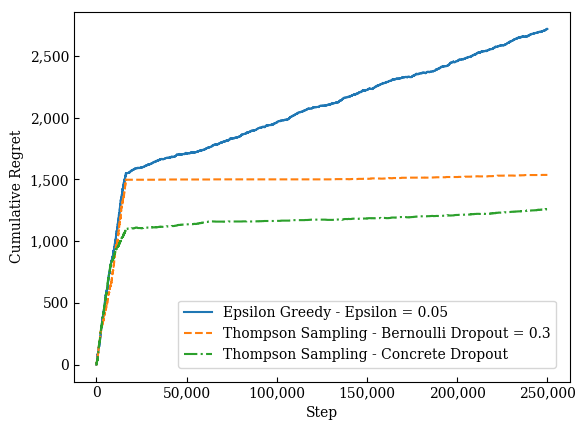}
\caption{Casino Parity Task - comparison between epsilon-greedy, Thompson sampling with Bernoulli dropout, and Thompson sampling with Concrete Dropout. Concrete Dropout with final cumulative regret (FCR) of 1,258 performs better than Bernoulli dropout which has FCR of 1,536 and the epsilon-greedy agent which has FCR of 2,718.}
\label{fig:casino}
\end{figure}

In \autoref{fig:casino} we see that Concrete Dropout with Thompson sampling, which is our proposed deep contextual multi-armed bandit, again outperforms fixed rate Bernoulli dropout and the epsilon-greedy agent. We note that for a non-contextual bandit the final cumulative regret was 25,102.

\section{CONCLUSION}

We have presented an approach to the contextual multi-armed bandit problem which enables modeling the context ($\mathbf{x}$) using a deep non-linear neural network while still enabling principled and efficient exploration of the joint context, action space $X \bigtimes A$.

We proposed deep contextual multi-armed bandits which apply recent work based on a Bayesian interpretation of dropout \cite{gal2016dropout,gal2017concrete}. By combining standard dropout training with inference time dropout, we are able to sample from an approximate posterior over the weights in a Bayesian neural network. This enables Thompson sampling, a heuristic which has been shown to be effective in addressing the exploration vs.\ exploitation trade-off.

Concrete Dropout, a continuous relaxation of Bernoulli dropout, has been shown to provide well-calibrated uncertainty estimates. In practical applications it is not possible to run many simulations to determine a good dropout annealing schedule. By applying Concrete Dropout in deep contextual multi-armed bandits we avoid having to define such a schedule for each task. Through standard gradient descent training we learn dropout rates which appropriately trade off exploration and exploitation. As we wish to deploy deep contextual multi-armed bandits to many different tasks which may require vastly different levels of exploration, adaptively learning the appropriate dropout rate is a key advantage of our approach over using standard Bernoulli dropout.

Deep contextual multi-armed bandits empirically outperform non-contextual bandits, bandits with epsilon-greedy exploration and fixed dropout rate bandits on the two contextual bandit tasks presented in this paper. Additionally we note that we are applying our approach to a number of contextual bandit problems in the marketing domain at HubSpot.

\subsubsection*{Acknowledgements}

The authors would like to thank Marco Lagi, Adam Starikiewicz, Vedant Misra and George Banis for their helpful comments on drafts of this paper.

\newpage


\bibliographystyle{splncs03}
\bibliography{references}

\begin{thebibliography}{10}
\providecommand{\url}[1]{\texttt{#1}}
\providecommand{\urlprefix}{URL }

\bibitem{agrawal2013thompson}
Agrawal, S., Goyal, N.: Thompson sampling for contextual bandits with linear
  payoffs. In: International Conference on Machine Learning. pp. 127--135
  (2013)

\bibitem{auer2002using}
Auer, P.: Using confidence bounds for exploitation-exploration trade-offs.
  Journal of Machine Learning Research  3(Nov),  397--422 (2002)

\bibitem{auer2002finite}
Auer, P., Cesa-Bianchi, N., Fischer, P.: Finite-time analysis of the multiarmed
  bandit problem. Machine learning  47(2-3),  235--256 (2002)

\bibitem{auer2002nonstochastic}
Auer, P., Cesa-Bianchi, N., Freund, Y., Schapire, R.E.: The nonstochastic
  multiarmed bandit problem. SIAM journal on computing  32(1),  48--77 (2002)

\bibitem{UCI2013}
Bache, K., Lichman, M.: UCI Machine Learning Repository

\bibitem{barber1998ensemble}
Barber, D., Bishop, C.M.: Ensemble learning in bayesian neural networks. NATO
  ASI SERIES F COMPUTER AND SYSTEMS SCIENCES  168,  215--238 (1998)

\bibitem{blundell2015weight}
Blundell, C., Cornebise, J., Kavukcuoglu, K., Wierstra, D.: Weight uncertainty
  in neural network. In: International Conference on Machine Learning. pp.
  1613--1622 (2015)

\bibitem{bubeck2012regret}
Bubeck, S., Cesa-Bianchi, N., et~al.: Regret analysis of stochastic and
  nonstochastic multi-armed bandit problems. Foundations and
  Trends{\textregistered} in Machine Learning  5(1),  1--122 (2012)

\bibitem{buntine1991bayesian}
Buntine, W.L., Weigend, A.S.: Bayesian back-propagation. Complex systems  5(6),
   603--643 (1991)

\bibitem{chapelle2011empirical}
Chapelle, O., Li, L.: An empirical evaluation of thompson sampling. In:
  Advances in neural information processing systems. pp. 2249--2257 (2011)

\bibitem{damianou2013deep}
Damianou, A., Lawrence, N.: Deep gaussian processes. In: Artificial
  Intelligence and Statistics. pp. 207--215 (2013)

\bibitem{gal2016dropout}
Gal, Y., Ghahramani, Z.: Dropout as a bayesian approximation: Representing
  model uncertainty in deep learning. In: International Conference on Machine
  Learning. pp. 1050--1059 (2016)

\bibitem{gal2017concrete}
Gal, Y., Hron, J., Kendall, A.: Concrete dropout. In: Advances in Neural
  Information Processing Systems. pp. 3584--3593 (2017)

\bibitem{graves2011practical}
Graves, A.: Practical variational inference for neural networks. In: Advances
  in Neural Information Processing Systems. pp. 2348--2356 (2011)

\bibitem{guez2015sample}
Guez, A.: Sample-Based Search Methods For Bayes-Adaptive Planning. Ph.D.
  thesis, UCL (University College London) (2015)

\bibitem{hinton1993keeping}
Hinton, G.E., Van~Camp, D.: Keeping the neural networks simple by minimizing
  the description length of the weights. In: Proceedings of the sixth annual
  conference on Computational learning theory. pp. 5--13. ACM (1993)

\bibitem{kingma2014adam}
Kingma, D.P., Ba, J.: Adam: A method for stochastic optimization. arXiv
  preprint arXiv:1412.6980  (2014)

\bibitem{li2010contextual}
Li, L., Chu, W., Langford, J., Schapire, R.E.: A contextual-bandit approach to
  personalized news article recommendation. In: Proceedings of the 19th
  international conference on World wide web. pp. 661--670. ACM (2010)

\bibitem{lu2010contextual}
Lu, T., P{\'a}l, D., P{\'a}l, M.: Contextual multi-armed bandits. In:
  Proceedings of the Thirteenth international conference on Artificial
  Intelligence and Statistics. pp. 485--492 (2010)

\bibitem{mackay1992practical}
MacKay, D.J.: A practical bayesian framework for backpropagation networks.
  Neural computation  4(3),  448--472 (1992)

\bibitem{mackay1995probable}
MacKay, D.J.: Probable networks and plausible predictions—a review of
  practical bayesian methods for supervised neural networks. Network:
  Computation in Neural Systems  6(3),  469--505 (1995)

\bibitem{maddison2016concrete}
Maddison, C.J., Mnih, A., Teh, Y.W.: The concrete distribution: A continuous
  relaxation of discrete random variables. arXiv preprint arXiv:1611.00712
  (2016)

\bibitem{minsky1969perceptrons}
Minsky, M., Papert, S.A.: Perceptrons: an introduction to computational
  geometry. MIT press (1969)

\bibitem{neal1995bayesian}
Neal, R.M.: Bayesian Learning for Neural Networks. Ph.D. thesis, University of
  Toronto (1995)

\bibitem{silver2016mastering}
Silver, D., Huang, A., Maddison, C.J., Guez, A., Sifre, L., Van Den~Driessche,
  G., Schrittwieser, J., Antonoglou, I., Panneershelvam, V., Lanctot, M.,
  et~al.: Mastering the game of go with deep neural networks and tree search.
  nature  529(7587),  484--489 (2016)

\bibitem{srivastava2014dropout}
Srivastava, N., Hinton, G., Krizhevsky, A., Sutskever, I., Salakhutdinov, R.:
  Dropout: A simple way to prevent neural networks from overfitting. The
  Journal of Machine Learning Research  15(1),  1929--1958 (2014)

\bibitem{thompson1933likelihood}
Thompson, W.R.: On the likelihood that one unknown probability exceeds another
  in view of the evidence of two samples. Biometrika  25(3/4),  285--294 (1933)

\end{thebibliography}

\end{document}